\documentclass{ieeetj}
\usepackage{cite}
\usepackage{amsmath,amssymb,amsfonts}
\usepackage{algpseudocode}  
\usepackage{graphicx,color}
\usepackage{subcaption}          
\usepackage{caption}
\usepackage{adjustbox}
\usepackage{textcomp}
\usepackage{xcolor}
\usepackage[table]{xcolor} 
\usepackage{bm}
\usepackage{pgfplots}
\usepackage{placeins}
\usepackage{pgfplotstable}
\usepackage{float} 
\usepackage{booktabs}
\usepackage{multirow}
\pgfplotsset{compat=1.17}
\usepackage{fix-cm}
\usepgfplotslibrary{groupplots}
\usepackage[hidelinks]{hyperref}
\newcommand{\doi}[1]{doi: \href{https://doi.org/#1}{\textcolor{blue!60!black}{#1}}}
\usepackage{algorithm}
\def\BibTeX{{\rm B\kern-.05em{\sc i\kern-.025em b}\kern-.08em
    T\kern-.1667em\lower.7ex\hbox{E}\kern-.125emX}}
\AtBeginDocument{\definecolor{tmlcncolor}{cmyk}{0.93,0.59,0.15,0.02}\definecolor{NavyBlue}{RGB}{0,86,125}}

\def\authorrefmark#1{\ensuremath{^{\textbf{#1}}}}
\begin{document}


\title{A Validation Strategy for Deep Learning Models:  Evaluating and Enhancing Robustness}
\author{Abdul-Rauf Nuhu\authorrefmark{1}, Parham Kebria\authorrefmark{1}, \IEEEmembership{Senior Member, IEEE}, Vahid Hemmati\authorrefmark{1}, Benjamin Lartey\authorrefmark{1}, Mahmoud Nabil Mahmoud\authorrefmark{2}, Abdollah Homaifar\authorrefmark{1}, \IEEEmembership{Senior Member, IEEE}, and Edward Tunstel\authorrefmark{3}, \IEEEmembership{Fellow, IEEE}
\affil{Department of Electrical and Computer Engineering, North Carolina A\&T State University, Greensboro, NC 27411 USA}
\affil{Department of Computer Science, University of Alabama, Tuscaloosa, AL 35487 USA}
\affil{Southwest Research Institute, San Antonio, TX 78238 USA}
\corresp{Corresponding author: Abdollah Homaifar (email: homaifar@ncat.edu).}
}
\begin{abstract}
Data-driven models, especially deep learning classifiers often demonstrate great success on clean datasets. Yet, they remain vulnerable to common data distortions such as adversarial and common corruption perturbations. These perturbations can significantly degrade performance, thereby challenging the overall reliability of the models. Traditional robustness validation typically relies on perturbed test datasets to assess and improve model performance.  In our framework, however, we propose a validation approach that extracts ``weak robust" samples directly from the training dataset via local robustness analysis. These samples, being the most susceptible to perturbations, serve as an early and sensitive indicator of the model’s vulnerabilities. By evaluating models on these challenging training instances, we gain a more nuanced understanding of its robustness, which informs targeted performance enhancement. We demonstrate the effectiveness of our approach on models trained with CIFAR-$10$, CIFAR-$100$, and ImageNet, highlighting how robustness validation guided by weak robust samples can drive meaningful improvements in model reliability under adversarial and common corruption scenarios.
\end{abstract}



\maketitle
\section{INTRODUCTION}
\IEEEPARstart{D}{ata-driven} models, especially deep learning models, recently achieved remarkable performance across a wide range of applications \cite{RaoufOJCS2025}. They play a crucial role in safety-critical applications \cite{10747123, 9659972}; however, they remain vulnerable to common data distortions, including adversarial attacks and common corruptions \cite{10806808, GUO2023109308}. For example, a standard AllConvNet (All Convolutional Network) model attains an error (top-$1$ error) of about $\textbf{6\%}$ on the CIFAR-$10$ test dataset, while on the adversarial and common corruption counterparts of the same dataset, its performance degraded significantly. On adversarial datasets and corruption datasets using basic iterative method (BIM) and Gaussian noise, the model attains an error of about $\textbf{80\%}$ and $\textbf{65\%}$, respectively. The safety concerns brought by these robustness-challenging scenarios have attracted attention towards developing procedures for model performance assessment \cite{GUO2023109308,8835375}, and robust models for adversarial datasets \cite{10316046, 10892039} or corruption datasets \cite{hendrycks*2020augmix,hendrycks2021many, 10767312}. Newly developed robust models are typically validated on datasets corresponding to their design focus, such as adversarial datasets for adversarial robustness and corruption datasets for common corruption robustness \cite{GUO2023109308}.  Robust techniques that work for adversarial datasets have shown to perform poorly on common corruption datasets and vice versa \cite{GUO2023109308, kireev2022effectiveness}. AugMix \cite{hendrycks*2020augmix} uses basic pre-defined corruptions to generate composed images to improve the robustness of models to common corruption datasets. DeepAugment \cite{hendrycks2021many}, generates corrupt images using distinct image-to-image models while altering the models’ parameters and activations with a wide array of manually defined heuristic operations. While both methods perform well for common corruption types in the CIFAR-10-C, CIFAR-$100$-C, and ImageNet-C datasets \cite{hendrycks2018benchmarking}, in comparison, they generalize poorly on adversarial datasets  \cite{calian2022defending}. The poor performance of different enhancement approaches on cross-domain datasets can be attributed to fewer domains included in data augmentation  operations \cite{gao2023out}. Targeted performance enhancement \cite{gao2023out, saikia2021improving, Hendrycks_2021_ICCV} have been suggested to include the most salient operations into the augmentation operations to enhance performance. This approach provides significant performance gains but is limited to corruption domains \cite{gao2023out}. We study the effectiveness of targeted augmentation on cross-domain of commonly studied perturbations (adversarial and common corruption types). To spot the salient operations, in our case, non-robust perturbation types, a model goes through an assessment on a wide array of adversarial and corruption datasets \cite{ 10062449, croce2021robustbench, carlini2019evaluating}.
\begin{figure*}[tbh]
\centering
\begin{subfigure}[b]{0.45\linewidth}
    \centering
    \includegraphics[width=0.95\linewidth]{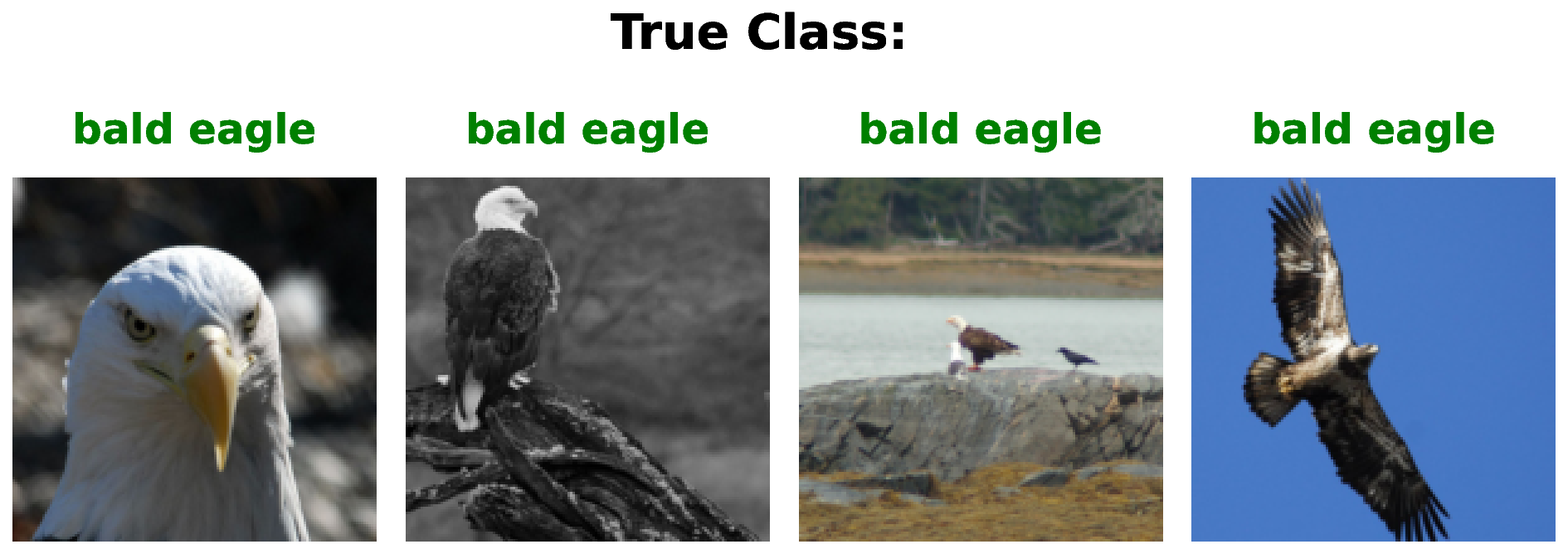}
    \caption{Original training data instances classified correctly as the true label "bald eagle."}
    \label{fig:weakRobust-a}
\end{subfigure}
\hspace{0.5cm}
\begin{subfigure}[b]{0.45\linewidth}
    \centering
    \includegraphics[width=0.95\linewidth]{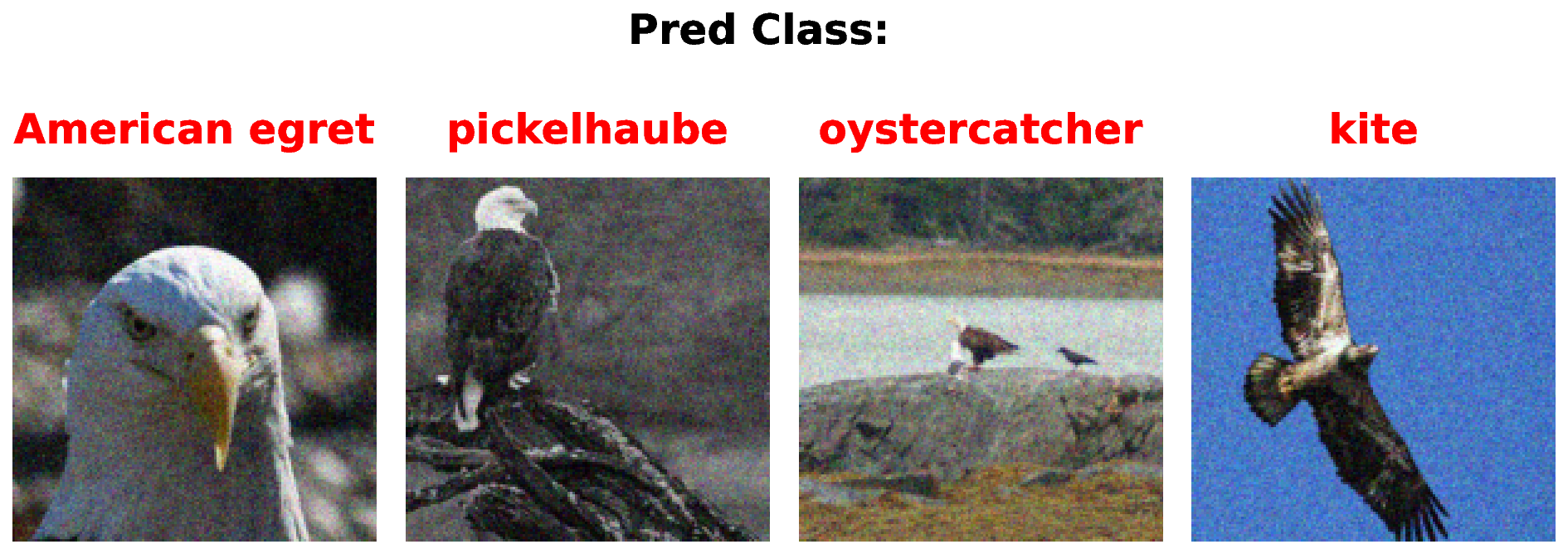}
    \caption{A perturbed version of the data instances in (a) misclassified to belong to different classes.}
    \label{fig:weakRobust-b}
\end{subfigure}
\vspace{-3mm}
\caption{A trained ResNet-18 correctly classifies the original instances (a) as bald eagle, while their perturbed variants with Gaussian noise of perturbation degree 0.1, imperceptible to the human eye, are misclassified. Instances whose majority of neighbors are misclassified are considered weak robust samples.}
\label{fig-weakRobust-Samples}
\end{figure*}
We explore how data augmentation enhances cross-domain generalization under the considered perturbation settings. First, we  analyze a standard model's performance on a validation dataset consisting of adversarial and corruption samples to benchmark non-robust perturbation types. Next, we generate an augmented dataset for adversarial and corruption categories using the corresponding non-robust perturbation types. Training on augmented and clean datasets significantly improves cross-domain performance and overall robustness \cite{gao2023out}. Unlike conventional robustness validation sets, our validation dataset comprises weak robust samples from the training dataset. We used weak robust samples as they are the most susceptible to perturbations and are an early indicator of models' vulnerabilities \cite{10316046, zhong2021understanding}. As shown in Fig. \ref{fig-weakRobust-Samples}, the original images (a) are weak robust samples, the model predicts the clean images as true class, bald eagle whereas as their perturbed variants, (b) are  misclassified by the model into different classes.
Weak robust samples can be identified from a given dataset via per-input (local) robustness analysis \cite{8103145}. Per-input robustness measures how consistently a model predicts perturbed inputs within a small neighborhood. The idea of using weak robust samples to evaluate deep learning models performance was first introduced in \cite{zhong2021understanding}. The authors aimed to address the limitations of existing robustness measurement methods \cite{engstrom2019exploring}, which rely on test datasets to assess model robustness. They noted traditional approaches estimate the overall robustness of models but fail to identify specific weaknesses. In \cite{zhong2021understanding}, the authors applied per-input robustness analysis to isolate weak robust samples in the input space. Unlike \cite{zhong2021understanding}, which uses weak robust samples merely to flag non-robust data points, we harness these samples to uncover robust and non-robust perturbation types under realistic conditions, such as weather
variations and norm-bound perturbations. etc.
A trained robust model is expected to perform well on corrupted versions of the training dataset from which it originated \cite{valentim2023evolutionary, xu2012robustness, zou2024towards}. Evaluating the model on a validation set composed of these corrupted samples can pinpoint the specific adversarial and common corruption types that expose its vulnerabilities \cite{papernot2016limitations, yang2019invariance}.

In a nutshell, we propose a novel validation procedure; It pinpoints the vulnerabilities of models through evaluation and improves their performance via targeted augmentation. Our validation procedure builds on the conventional holdout validation (hoVa) approach \cite{7497471}, but it takes a distinct path and is different in several ways.

First, we assess robustness using weak robust samples rather than evaluating model performance on a reserved clean test set. These are training data points the model has seen but still struggles with under minor perturbations when testing its performance. Our approach reveals how well a model handles slight variations even in familiar data.

Second, deliberately perturbing the weak robust samples, our method uncovers subtle weaknesses that are not captured via hoVa. This targeted evaluation mirrors real-world challenges, such as noise and weather distortions. It enables early detection of vulnerabilities and guides targeted robustness enhancements of models. In doing so, our approach offers a more realistic assessment of models' performance under adverse conditions.

Moreover, our procedure is especially beneficial in data-scarce scenarios. Perturbing weak robust samples with various perturbations, we create an expanded and diverse evaluation dataset. This strategy allows models to be trained on the entire training dataset, while their performance is rigorously assessed on a rich set of challenging instances. This ensures robust evaluation even when test data is limited \cite{shorten2019survey}.

We enhance model robustness across the cross-domain scenarios by identifying the perturbation types to which the model is most vulnerable. These weaknesses are leveraged during retraining. The process combines data augmentation with an updated loss function. The augmented dataset includes original samples, adversarial counterparts, and common corruption counterparts. To create the adversarial counterparts, we use one of the adversarial perturbation types that exposes the models' weaknesses. Similarly, to create common corruption counterparts, we update the default AugMix \cite{hendrycks*2020augmix} augmentations. Within each category, we selectively incorporate the corruption type whose corresponding validation dataset yields the highest model error. Unlike AugMix, our method is a targeted augmentation approach that selects the most challenging perturbation types through a validation-driven procedure. This ensures that augmentations specifically target the model’s weaknesses. By doing so, our approach improves both adversarial and common corruption robustness more effectively than existing techniques. We refer to this strategy as \textbf{R}obustness \textbf{E}nhancement via \textbf{Va}lidation (REVa).


In summary, the major contributions of this paper are as follows: 

\begin{enumerate}
    \item \textit{Propose per-input resilient analyzer, a procedure that measures the robustness properties of deep learning models and guides rearranging a given training dataset in order of the weakest robust to strong robust samples. This procedure provides an effective way of benchmarking the models robustness under different degrees of samples robustness. A validation dataset derived from weak robust samples of CIFAR and ImageNet training datasets is proposed. We made our dataset preparation procedure publicly available.\footnote{\url{https://github.com/arnuhu08/REVa} \label{repo}}}
    
    \item \textit{Development of an evaluation procedure that employs different evaluation metrics to assess the robustness of models using the validation dataset. It uses two distinct metrics tailored to categorize adversarial and common corruption types into robust and non-robust groups. This validation procedure yields valuable insights into the strengths and weaknesses of models, serving as a foundation for guiding robustness improvements.}
    
    \item \textit{Propose REVa, a procedure designed to enhance a model robustness by targeting identified non-robust perturbation types. Through targeted augmentation, REVa improves model performance. We generate an augmented training dataset by incorporating these non-robust perturbations, and retrain the model using an updated loss function. This approach significantly enhances robustness across various adversarial and corruption scenarios. Comparative results demonstrate that our method outperforms existing techniques, yielding substantial performance improvements}

\end{enumerate}

The remainder of the paper is organized as follows. Section \ref{section:2} provides some background and preliminaries of the problem formulation. The proposed validation approach is discussed in Section \ref{section:3}. Section \ref{section:4} presents the outcome of the validation procedure and performance of its end product, the robustness enhancement strategy. Finally, Section \ref{section:5} concludes the paper and outlines future work.

\section{Preliminary}\label{section:2}
This section presents the terminologies used in this paper and the necessary background information.

\subsection{Terminology Definitions}
We provide a brief discussion of  key terminologies used.\\

\begin{figure}[tbh]
\centering
\includegraphics[width=0.8\linewidth]{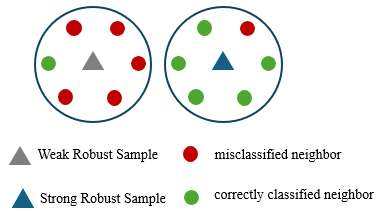}
\vspace{\floatsep}
\caption{Data instances and the corresponding neighbors. The gray and blue triangles show the original instances; the green and red circles are the neighbors generated from perturbing an original instance.}
\label{fig-ngbs}
\end{figure}

\textbf{Neighbors:} \textit{Samples generated by perturbing the original instances of a given dataset.} For example, in Fig. \ref{fig-weakRobust-Samples}, images in (a) are the original instances, and their corresponding neighbors are in (b). Neighbors can be generated using natural variation or using random perturbation techniques. We provide a clear definition of neighbor in Fig.  \ref{fig-ngbs}. The gray and blue triangles are original instances whereas the green and red circles represent their neighbors. These neighbors are generated via random perturbation.

\par \textbf{Weak Robust Samples.} 
A data instance is considered a \emph{weak robust sample} if the model misclassifies many or all of its neighbors under small perturbations, indicating low stability of the decision boundary. As illustrated in Fig.~\ref{fig-weakRobust-Samples}, the original images in (a) are correctly classified by the model, whereas their perturbed neighbors in (b) are misclassified, demonstrating weak robustness. This concept is theoretically supported by margin-based stability frameworks \cite{NEURIPS2024_29753d93, xu2012robustness}, which show that samples with small classification margins or fragile neighborhoods significantly affect robustness. Consequently, weak robust samples provide practical indicators of margin-sensitive regions in the training set, even when clean training accuracy is close to $100\%$ \cite{NEURIPS2024_29753d93}.

\vspace{-8mm}
\begin{figure*}[t]
\centering
\includegraphics[width=45pc]{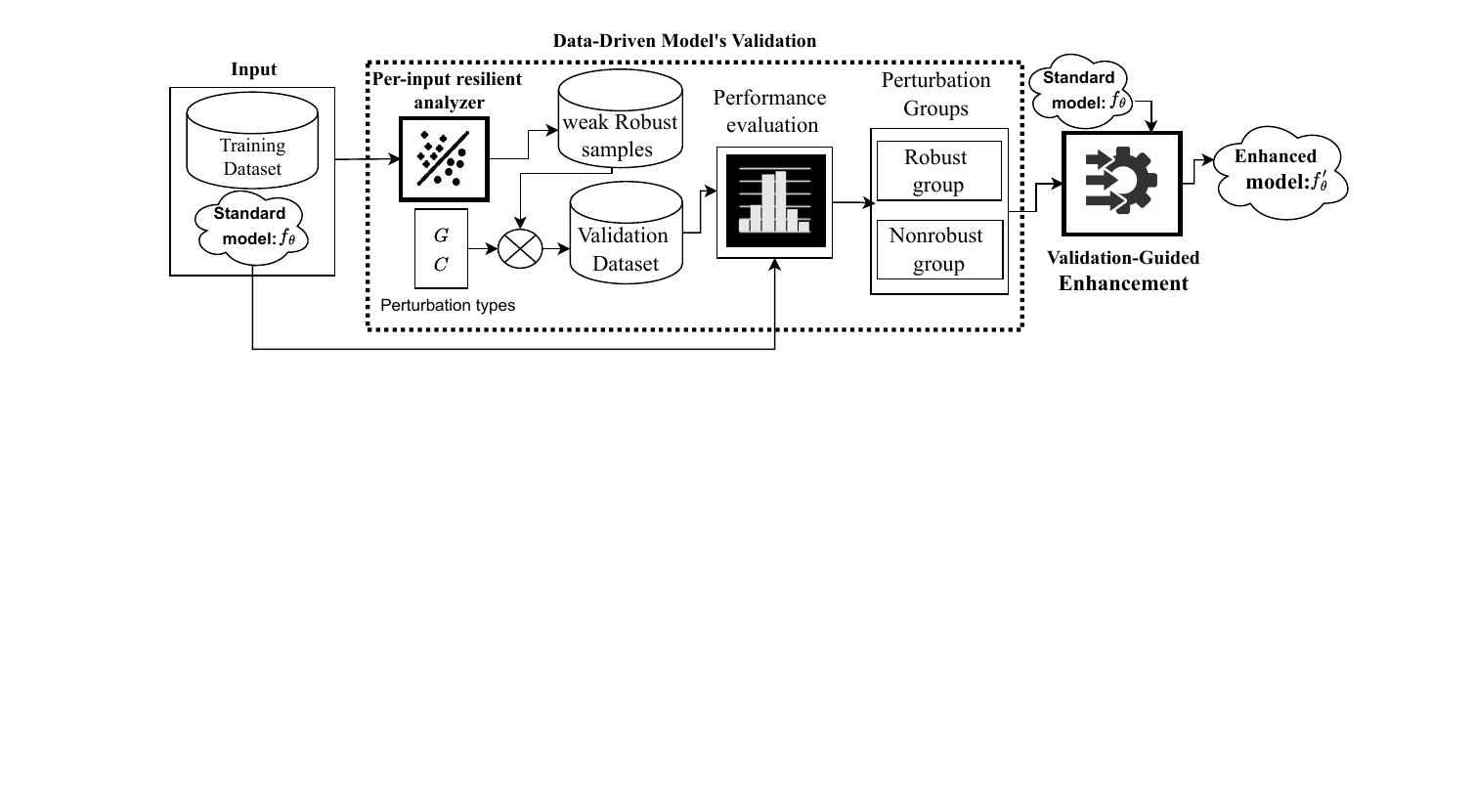}
\vspace*{-60mm}
\caption{ Overview of the proposed framework. Starting with a standard trained model $f_\theta$
and a training dataset, the Per-input resilient analyzer identifies weak robust samples through local robustness analysis. These samples are perturbed with a range of distortions -- both adversarial ($G$) and common corruption ($C$) -- to create a validation dataset for an in-depth robustness assessment. Based on model performance, the perturbation types are then categorized into robust and non-robust groups. Finally, model 
 $f'_\theta$ is developed by refining  $f_\theta$ using the non-robust perturbations, enhancing its robustness. } 
\label{fig:1-proposed framework}
\end{figure*}
\subsection{Problem Description and Formulation} 
\par \textbf{Problem Description:} 
Given a deep learning classifier that performs well on clean test data, our goal is to elevate its robustness against a given cross-domain (adversarial and common corruption perturbations) via targeted enhancement. We begin by reorganizing the clean training dataset, ranking training instances from least robust to most robust. By evaluating the classifier on perturbed versions of the weakest samples, we identify the vulnerabilities of both perturbation types that hinder performance. We use the non-robust perturbations to guide our enhancement process,  ultimately producing a reliable model in the presence of adversarial and common corruption datasets.\\
\par \textbf{Problem Formulation}\\
In the remainder of this paper: 
 \begin{itemize}
     \item \bm{$f_{\theta}$} denotes a function of a deep learning classifier parameterized by its weights $\theta$. 
     \item \bm{$f'_{\theta}$} denotes an enhanced version of \bm{$f_{\theta}$}. 
     \item  \bm{$\mathcal{X} = \{x_1, x_2, \ldots, x_n \}$} with \bm{$x_i\in\mathbb{R}^{d}$}, represents the clean training dataset of size \bm{$n$} and \bm{$\mathcal{Y} = \{y_1, y_2, \ldots, y_m \}$} with \bm{$y_i\in\mathbb{N}$}, represents \bm{$m$} distinct class labels. Where, $d=H\times W\times C$ is the size of RGB image (with C=3) .
     \item  \bm{$\mathcal{Z} = \{z_1, z_2, \ldots, z_r \}$} with \bm{$z_i\in\mathbb{R}^{d}$}, represents clean test dataset of size \bm{$r$} and the same target labels as \bm{$\mathcal{X}$}.
     \item \bm{$ G = \{g_1, g_2, \ldots, g_k\}$} with \bm{$k\in \mathbb{N}$}, represents a set of adversarial generation techniques.
     \item \bm{$ C = \{c_1,\ldots, c_l\}$}  with \bm{$l\in \mathbb{N}$}, represents a set of common corruption types.
 \end{itemize}
 Consider \bm{$\mathcal{T}$} as a performance satisfaction score of the classifier \bm{$f_\theta$} on the clean test dataset \bm{$\mathcal{Z}$}, such that \bm{$f_\theta(\mathcal{Z}) \geq \mathcal{T}$}. For instance, \bm{$\mathcal{T}$}  can be a test accuracy threshold. There exists a procedure \bm{$\mathcal{P}$} that assesses the local robustness of a data instance $x_i$ on \bm{$f_\theta$}  and assigns a robust confidence, \bm{$\gamma_i$} to that instance. \bm{$\gamma_i$} quantifies the confidence of the model \bm{$f_{\theta}$} when evaluated on the input $x_i$. In our case, \bm{$\gamma_i$} is the average of the highest confidence scores \bm{$f_{\theta}$} assigned to the neighbors of $x_i$ when they are misclassified. Given \bm{$\mathcal{S} = \{x_1, x_2, \ldots, x_w \} \subset \mathcal{X}$}, that represents weak robust samples from \bm{$\mathcal{X}$}. \bm{$\mathcal{S}$} contains equal samples of each class in \bm{$\mathcal{Y}$}. Knowing \bm{$\mathcal{S}$}, we generate \bm{$\mathcal{S}^{(g)}:g\in G$} and \bm{$\mathcal{S}^{(c)}:c\in C$}. We denote the collection of adversarial datasets generated from \bm{$\mathcal{S}$} as \bm{$\mathcal{S}^{G}_{adv} = \{\mathcal{S}^{(g)}:g\in G \}$}. Similarly, \bm{$\mathcal{S}_{\mathrm{cc}}^{C}= \{ \mathcal{S}^{(c)}:c\in C\}$}, denotes a collection of common corruption datasets generated from \bm{$\mathcal{S}$}. We propose an approach that evaluates and enhances the performance of models on datasets perturbed by \bm{$G$} and \bm{$C$}.
 \par To tackle this problem, we decompose it into three sub-problems as listed below.
 \begin{enumerate}
     \item [] \textbf{Sub-Problem 1:} Given \bm{$f_\theta$} and \bm{$\mathcal{X}$}, extract the least robust samples \bm{$\mathcal{S}$} from \bm{$\mathcal{X}$} based on local robustness analysis of \bm{$f_\theta$} for generating \bm{$ \mathcal{S}_{\mathrm{adv}}^{G}$} and \bm{$\mathcal{S}_{\mathrm{cc}}^{C}$}. 
     \item [] \textbf{Sub-Problem 2:} Given \bm{$ \mathcal{S}_{\mathrm{adv}}^{G}$} and \bm{$ \mathcal{S}_{\mathrm{cc}}^{C}$} assess \bm{$f_\theta$} on dataset corresponding to each perturbation type. Then based on \bm{$f_\theta$} performance, classify the perturbation types in \bm{$G$} and \bm{$C$} into robust group, $\rho$ or non-robust group, $\nu$. 
    \item [] \textbf{Sub-Problem 3:} Given the non-robust group $\nu$, develop a robust model \bm{$f_{\theta}'$} that results in improved performance on datasets perturbed using both $\rho$ and $\nu$ groups.
 \end{enumerate}
This formulation outlines our approach of using weak robust samples from the training dataset for targeted perturbation analysis. It also highlights the utilization of the outcomes of targeted perturbation analysis to guide the robustness enhancement across a considered spectrum of perturbation types.
\section{Methodology}\label{section:3}
In this section, we explain the proposed framework in details. 
Fig. \ref{fig:1-proposed framework} illustrates the framework, which takes a
trained model along with the dataset it was trained on as input. The framework outputs an enhanced version of the given model and consists of two key components: validation and
validation-guided model enhancement.
\begin{algorithm}[tbh]
    \setlength{\textfloatsep}{5pt}  
    \small 
    \caption{Per-input Resilient Analyzer}
    \label{algo:1}
    \begin{algorithmic}[1]
        \Require \(f_\theta\), \(\mathcal{X}\), \(\mathcal{Y}\), \(\kappa\), \(\epsilon\)
        \Ensure Dictionary \(W\) mapping each training instance \(x_i\) to misprediction score \(\gamma_i\), sorted in descending order
        
        \Function{ComputeMispredictionScore}{$x$, $y$}
            \State \(neighbor\_sum \gets 0\)  
            \For{\(q = 1\) \textbf{to} \(\kappa\)}
                \State \(x_q \gets x + \epsilon \cdot \text{random}(\text{size}(x))\)  
                \State \(neighbor\_sum \gets neighbor\_sum + M_s(x_q)\)

            \EndFor
            \State \Return \(\frac{neighbor\_sum}{\kappa}\)
        \EndFunction
        
        \State \(W \gets \{\}\)  
        \For{each class \(y_i \in \mathcal{Y}\)}
            \State Initialize list \(R \gets []\)  
            \State Find \(v_{c}\): samples in \(\mathcal{X}\) belonging to class \(y_i\)  
            \For{\(i = 1\) \textbf{to} \(|v_{c}|\)}
                \State \(\gamma_i \gets \Call{ComputeMispredictionScore}{x_i, y_i}\)  
                \State Append \((x_i, \gamma_i)\) to \(R\)
            \EndFor
            \State \textbf{Sort} \(R\) in descending order by \(\gamma_i\)  
            \State \(W[y_i] \gets R\)
        \EndFor
        \State \Return \(W\)
    \end{algorithmic}
\end{algorithm}

\subsection{\textbf{Input Model Validation}} It consists of two main parts: The per-input resilient analyzer and the performance evaluation of the input model. This component aims to explore strengths($\rho$) and weaknesses($\nu$) of a model.
\vspace{-5mm}
\subsubsection{Per-input resilient analyzer} 
Algorithm \ref{algo:1} summarizes the per-input resilient analyzer. It orders samples from the least robust to the most robust. It takes a trained model \bm{$f_\theta$} and its training dataset \bm{$\mathcal{X}$} as inputs. First, an empty dictionary $W$ is created to store each class's sorted list of samples. Each instance
$x_i$ in each class goes through $\kappa$ random sampling to
generate a set of $\kappa$ perturbed neighbors. Every neighbor in $\kappa$ is bounded by a perturbation degree, $\epsilon$. Then, each perturbed data instance $x_q, q\in\kappa$ is passed through \bm{$f_\theta$} to obtain the probability of possible output labels. If a neighbor is classified correctly, a misprediction score ($M_s$) of $0$ is assigned to that neighbor. On the other hand, if a neighbor is misclassified, its $M_s$ is assigned as the maximum probability. This misprediction schema is presented in Equation \eqref{misprediction}. A score, $\gamma_i$ is assigned to each data instance $x_i$ by taking the average of its neighbors' misprediction scores ($M_s$). After assigning a score to all instances in a class, the samples are stored in descending order with respect to $\gamma_i$; from the least robust samples to the most robust samples. Finally, $W$ is returned as a dictionary that maps each class to its corresponding sorted list of samples.\\ 

\begin{equation}
M_s(x_q) =
\begin{cases}
0, & \text{if } \hat{y}_q = y_i\\
\max_{k \neq y_i} \text{softmax}(z(x_q))_k, & otherwise
\end{cases}
\label{misprediction}
\end{equation}
\subsubsection{Performance evaluation of the input model}
First,  we construct a validation dataset composed of adversarial and common corruption samples derived from the weak robust portion of the training dataset. Weak robust samples are chosen because models remain vulnerable to variations such as random perturbation when classifying them. They represent worst-case clean instances that models may encounter in real-world applications. This makes them ideal for robustness testing \cite{zhong2021understanding, zhao2024attack}. To select weak robust samples, we use the output of the per-input resilient analyzer, $W$. The samples are selected from $W$ in such a way that their size matches the size of the original test dataset. For each class, $y$ we select the first $\Psi$ samples from that class. Where $\Psi$ is a user-preferential number of samples per class. We generate the validation dataset by applying a variety of adversarial and corruption perturbations to the selected weak robust samples. Specifically, we used $6$ different adversarial and $15$ common corruption perturbations. The adversarial techniques used include FGSM (Fast Gradient Sign Method)\cite{DBLP:journals/corr/GoodfellowSS14}, PGD (Projected $L_{\infty}$ Gradient Descent)\cite{madry2018towards}, BIM (Basic Iterative Method)\cite{kurakin2018adversarial}, RFGSM (Random perturbation with FGSM)\cite{tramer2018ensemble}, UMIFGSM (Untargetted Momentum Iterative FGSM)\cite{dong2018boosting}, and UAP (Universal Adversarial Perturbation)\cite{moosavi2017universal}. Similarly, the common corruption perturbations are sourced from standard benchmarks \cite{hendrycks2018benchmarking}. These corruptions are categorized into weather, digital, blur, and noise. We focus on untargeted adversarial attacks, which are assumed to occur naturally \cite{levy2022roma}. Unlike existing validation datasets, our dataset clearly distinguishes between perturbation types that challenge model robustness.  Once the validation dataset is created, a post-training performance evaluation is then done. We assess the performance of standard trained models across the validation dataset's adversarial and common corruption versions. Using Equations \eqref{eqn_3rE} and \eqref{eqn_4rE},  a model's robustness is assessed. For both adversarial and  corruption datasets, we categorize each type into robust and non-robust groups. The robust group includes perturbation types when used to perturb weak robust samples, models performance do not degrade significantly when assessed on these perturbed datasets. In contrast, the non-robust group consists of perturbation types when used to perturb weak robust samples cause a substantial drop in the models performance when evaluated on them. The robust groups highlight a model's strengths whereas the non-robust groups highlight a model's weaknesses in terms of the perturbation types.
\subsection{Validation-Guided Enhancement of Models}
We employ targeted augmentation techniques to train robust models. Models are trained on augmented datasets drawn from adversarial and common corruption settings. The augmented dataset consists of the original training dataset, its adversarial counterparts, and common corruption counterparts. We used PGD technique \cite{madry2018towards} from the non-robust groups to generate the adversarial counterparts. This technique shows better performance improvement than other adversarial techniques for the adversarial counterparts of the training dataset generation. Similarly, AugMix \cite{hendrycks*2020augmix} with updated augmentation operations is used to generate the common corruption counterparts. We update
the AugMix default augmentation operations with
each category's most challenging common corruption types. Since each default operation has an associated severity level. We use the severity level of that common corruption type on which the model starts performing badly.\\
\par \textbf{Updated Loss Function:} We update the training loss function with a modified loss function to ensure the models exhibit smooth responses across the different datasets. Given that the semantic information in the adversarial and  common corruption datasets are still approximately preserved, the model should process the original, adversarial, and  common corruption datasets similarly \cite{hendrycks*2020augmix}. To achieve this, we incorporated the Jensen-Shannon Divergence Consistency loss (JS), as proposed in the AugMix paper \cite{hendrycks*2020augmix}, and introduced an adversarial loss. The adversarial loss ensures a smoother response to adversarial examples, while the Jensen-Shannon loss promotes a smoother response to the  common corruption dataset. The original training loss is then replaced by the loss as expressed in Equation \eqref{eqn_loss}.
\begin{equation}
\begin{aligned}
    \mathcal{L}(p_{orig}, y) + \beta \cdot JS(p_{orig}, p_{augmix})
    + \alpha \cdot \mathcal{L}(p_{adv}, y)
\end{aligned}
\label{eqn_loss}
\end{equation}

Where $p_{orig} = \hat{p}(y \; \mid \; x_{orig})$ represents the probability distribution of the original sample, $x_{orig}$. And $p_{augmix} = \hat{p}(y \; \mid \; x_{augmix})$, and $p_{adv} = \hat{p}(y \; \mid \; x_{adv})$ represent the probability distributions of $x_{orig}$ common corruption variant, $x_{augmix}$ and its adversarial variant, $x_{adv}$, respectively. $\alpha$ is a priority weight that indicates the weight assigned to the adversarial and  common corruption datasets. $\beta$ is defined as ($1$ - $\alpha$)$\cdot\lambda$, where $\lambda$ is the default regularization parameter in \cite{hendrycks*2020augmix}. Fine-tuning $\alpha$ improves robustness against adversarial and common corruption examples without sacrificing accuracy on clean data performance.
The updated loss function compels the model to maintain stability, consistency, and insensitivity across the original, adversarial, and  common corruption datasets. Our enhancement procedure (REVa) trains robust models by utilizing the augmented dataset and the updated loss function. 
\section{Experimental Study} \label{section:4}
In this section, experiments are conducted on well-known datasets. We provide a brief comparison of the results of the evaluation of models on our validation dataset compared with the existing validation dataset. Also, we present the comparison results of models enhanced via our proposed enhancement procedure (\textbf{REVa}) with standard models and their AugMix-enhanced versions.\\
\vspace{-5mm}
\subsection{Experimental setup} 
\subsubsection{\textbf{Dataset}} We considered three publicly available datasets: CIFAR (CIFAR-$10$, CIFAR-$100$) \cite{krizhevsky2009learning} and ImageNet \cite{ILSVRC15}. Both CIFAR datasets contain 32 x 32 RGB images with $50,000$ training and $10,000$ testing images. CIFAR-$10$ has 10 classes whereas CIFAR-$100$ has $100$ classes. The ImageNet contains 224 x 224 RGB images of approximately $1.2$ million training and $50,000$ testing instances categorized into $1000$ classes. For efficiency and timely manners, we used a subset of the ImageNet dataset, ImageNet-$100$ \cite{tian2020contrastive}. ImageNet-$100$ consists of approximately $120,000$ training and $5,000$ testing instances categorized into $100$ classes.\\

\textbf{Our validation dataset:} We denote adversarially perturbed datasets of our validation dataset as CIFAR-$10$-adv$^1$, CIFAR-$100$-adv$^1$, and ImageNet-$100$-adv$^1$. For the common corruption-perturbed datasets, we denote them as CIFAR-$10$-C$^1$, CIFAR-$100$-C$^1$, and ImageNet-$100$-C$^1$. We then generate our validation dataset as follows. For the CIFAR adversarial datasets (CIFAR-$10$-adv$^1$ and CIFAR-$100$-adv$^1$), we first select $10,000$ samples with equal distribution from each class of the reordered training data. Then, for each adversarial perturbation type, we create an adversarially perturbed version of the selected dataset. For ImageNet-$100$-adv$^1$,  we apply the same procedure as for the CIFAR adversarial datasets, but we select $5,000$ samples to match the test set size. Similarly, for the common corruption datasets (CIFAR‑10‑C¹, CIFAR‑100‑C¹, ImageNet‑100‑C¹), we use the same selected samples and perturb each with various common corruption types at 5 severity levels, following \cite{hendrycks2018benchmarking}.\\

\textbf{Performance comparison datasets:} We compare the performance of models enhanced via different approaches, we evaluate on the following datasets: CIFAR-$10$-adv, CIFAR-$100$-adv, ImageNet-$100$-adv, CIFAR-$10$-C, CIFAR-$100$-C and ImageNet-$100$-C. The "adv" datasets are the adversarial versions of the respective CIFAR and ImageNet-$100$ holdout test datasets using the adversarial techniques. Similarly, the "C" datasets are the corrupted versions of the respective CIFAR and ImageNet-$100$ holdout test datasets using the common corruption types in  \cite{hendrycks2018benchmarking}.
\vspace{-5mm}
\subsubsection{\textbf{Evaluation metrics}} We explain the evaluation metrics used for assessing the robustness of the models and performance comparison of different enhancement techniques. 

To categorize the adversarial perturbation types into robust and non-robust groups, we define relative adversarial error ($\mathrm{RAdv_{Err}}$). It measures how the performance of a model deteriorates when evaluated on adversarial perturbed dataset compared to its baseline error on the clean test dataset. The higher this value, the more vulnerable the model is to the corresponding adversarial dataset. Thus, the adversarial technique will be classified in the non-robust group. $\mathrm{RAdv_{Err}}$ is presented in Equation \eqref{eqn_3rE} as follows: 
\begin{equation}
    \mathrm{RAdv_{Err}}(f_\theta, \mathbb{D}, g_{\epsilon}) = \frac{\mathrm{Adv_{Err}}(f_\theta, \mathbb{D}, g_{\epsilon})}{\mathrm{Err}}
    \label{eqn_3rE}
\end{equation}

Where $f_\theta$ denotes the standard model, $\mathbb{D}$ a given dataset, and $\mathrm{Adv_{Err}}$, adversarial error is defined in Equation \eqref{eqn_3}. Also, $g_{\epsilon}$ denotes an adversarial technique with perturbation degree, $\epsilon$, and Err is the clean top-$1$-error rate.\\
Similarly, to categorize the common corruption types into robust and non-robust groups, we define relative corruption error ($\mathrm{RC_{Err}}$). Just like $\mathrm{RAdv_{Err}}$, it measures a model's performance deterioration on common corruption perturbed dataset relative to its baseline error on the clean test dataset. $\mathrm{RC_{Err}}$ is presented in Equation \eqref{eqn_4rE} as follows:
\begin{equation}
    \mathrm{RC_{Err}}(f_\theta, \mathbb{D}, c) = \frac{\mathrm{uCE}(f_\theta, \mathbb{D}, c)}{\mathrm{Err}}
    \label{eqn_4rE}
\end{equation}

Where $c$ denotes a corruption type, $\mathrm{uCE}$, unnormalized corruption error defined in Equation \eqref{eqn_4}. All other symbols are the same as those in Equation \eqref{eqn_3rE}.\\

To evaluate the performance of the enhanced models on adversarial and common corruption perturbation datasets, we report the errors and their mean. On the adversarial perturbation datasets, we report the adversarial error ($\mathrm{Adv_{Err}}$) for an adversarial technique in Equation \eqref{eqn_3} as follows:
\begin{equation}
    \mathrm{Adv}_{\mathrm{Err}}(f'_\theta, \mathbb{D}, g_{\epsilon}) = 
    \frac{1}{|\mathbb{D}|} \sum_{i=1}^{|\mathbb{D}|} 
    \mathbb{I}\big(f'_\theta(g_{\epsilon}(x_{i})) \neq y_{i}\big)
    \label{eqn_3}
\end{equation}
Where $f'_{\theta}$ denotes an enhanced model, $\mathbb{I}$ an indicator function, and All other symbols are the same as those in Equation \eqref{eqn_3rE}.\\

Besides the adversarial error, we report the mean adversarial error in Equation \eqref{eqn_3E} as follows: 

\begin{equation}
    \mathrm{mAdv}_{\mathrm{Err}}(f'_\theta, \mathbb{D}) = \frac{1}{|\mathbb{G}|} \sum_{k=1}^{|\mathbb{G}|} \mathrm{Adv}_{\mathrm{Err}}(f'_\theta, \mathbb{D}, g_{\epsilon,k})
    \label{eqn_3E}
\end{equation}

Where $|\mathbb{G}|$ denotes the total number of the adversarial techniques, and  $g_{\epsilon, k}$ the $k^{th}$ adversarial technique with a perturbation degree, $\epsilon$. \\

Like adversarial datasets, on common corruption datasets, we report the unnormalized error ($\mathrm{uCE}$). This involves first computing the error values for a common corruption type $c$ at different severity levels, $\mathbb{S}$. These error values are then averaged across the total severity levels to compute $\mathrm{uCE}$. $\mathrm{uCE}$ is presented in Equation \eqref{eqn_4} as follows:    
\begin{equation}
\begin{split}
    \mathrm{uCE}(f'_\theta, \mathbb{D}, c) = \\ 
    \frac{1}{|\mathbb{L}|} \sum_{l=1}^{|\mathbb{L}|} 
    \Bigg(
        \frac{1}{|\mathbb{D}|} \sum_{i=1}^{|\mathbb{D}|} 
        \mathbb{I}(f'_\theta(c_l(x_i)) \neq y_i)
    \Bigg)
\end{split}
\label{eqn_4}
\end{equation}

Where $\mathbb{L}$ denotes the number of severity levels and $c_{l}$ a corruption type at given severity degree, $l$.\\

Besides the $\mathrm{uCE}$, we also report on the mean corruption error ($\mathrm{mCE}$). We averaged over all the corruption types $\mathrm{uCE}$ values. $\mathrm{mCE}$ is presented in Equation \eqref{eqn_4E} as follows: 

\begin{equation}
     \mathrm{mCE} (f'_\theta, \mathbb{D}) = \frac{1}{\mathbb{|C|}}\sum_{j=1}^{\mathbb{|C|}}\Bigg(\mathrm{uCE}(f'_\theta, \mathbb{D}, c)\Bigg)
    \label{eqn_4E}
\end{equation}
Where $|\mathbb{C}|$ is the total number of corruption types considered and all other symbols are the same as those in Equation \eqref{eqn_4}.

\subsection{Implementation details}\label{implement:details} 
\par \textbf{Baseline Methods:} We compare the effectiveness of our proposed enhancement procedure with two baseline methods; standard and AugMix. We choose AugMix \cite{hendrycks*2020augmix} for comparison as it outperforms standard adversarial training (AT) \cite{madry2018towards} in robustness to common corruptions. This makes it a strong enhancement approach for performance comparison to our approach, which enhances robustness against both adversarial and corruption settings. Standard denotes a model trained only on the original CIFAR and ImageNet-$100$ datasets, whereas AugMix denotes models enhanced using the AugMix strategy. Also, we observed the effectiveness of our validation procedure in the proposed framework shown in Fig. \ref{fig:1-proposed framework} with REVa$^-$. REVa$^-$ mirrors our proposed enhancement method, REVa. It only differs in its use of AugMix's default augmentation operations to generate the corruption counterparts of the training dataset. Essentially, REVa$^-$ represents an AugMix-style training strategy with a slight modification. It incorporates an adversarial dataset component informed by our validation procedure, along with an extra adversarial loss term in the training objective.\\
\par \textbf{Training Setup:} To ensure a fair comparison of our targeted resilient enhancement training procedure, we utilized model architectures commonly employed in image classification tasks. Following the baseline resilient training procedure AugMix, we select the following deep learning architectures for our comparison. The considered models for the CIFAR dataset are: All Convolutional Network \cite{springenberg2014striving}, a DenseNet-BC \cite{huang2017densely}, and a 40-2 Wide ResNet \cite{zagoruyko2016wide}. For ImageNet-$100$, we conducted experiments using ResNet-$18$ \cite{he2016identity} and Swin\_V2\_B, a transformer model \cite{Liu_2022_CVPR}. We selected Swin V2 B for its high clean accuracy and computational efficiency on ImageNet-$1$K. This balance of accuracy and efficiency makes it a strong baseline for robustness evaluation.  For CNN-based models, we followed the training setup and parameters provided in \cite{hendrycks*2020augmix}. For Swin\_V2\_B, we used the default PyTorch parameters but excluded strong augmentations (e.g., Mixup and CutMix) to ensure a fair comparison with baseline methods. Both ResNet-$18$ and Swin\_V2\_B were trained for $100$ epochs. To adapt Swin\_V2\_B efficiently, we used progressive layer unfreezing for transfer learning. At first, only the classification head was trained while the backbone remained frozen. At epoch 5, all layers were unfrozen for full fine-tuning.

To give equal priority for the models on the adversarial and common corruption datasets, we set the value of $\alpha$ in Equation \eqref{eqn_loss} as $0.5$. For $\beta = (1 - \alpha)\cdot\lambda$, we set the value of $\lambda$ as provided in \cite{hendrycks*2020augmix}. 

For the Per-input resilient analyzer, the perturbation magnitude $\epsilon$ was determined via sensitivity analysis, assessing model accuracy across a range of values. The optimal $\epsilon$ corresponded to approximately $50\%$ accuracy, striking a balance between robustness evaluation and perceptual imperceptibility. Across datasets, $\epsilon$ varied from $0.20$ to $0.22$, and we fixed $\epsilon = 0.22$ for consistency. Additionally, we set the parameter $\kappa$ to $50$ following prior work \cite{zhong2021understanding}, which demonstrated that the average misprediction score stabilizes at this value and for large $\kappa$. This choice ensures reliable results while avoiding extra computational overhead.

To generate the adversarial training datasets, we used $\text{PGD}_{\infty}$. For CIFAR, the perturbation magnitude was set to $\epsilon = 8/255$, with a step size $\alpha = 2/255$ and 40 iterative steps. For ImageNet-$100$, we used $\epsilon = 4/255$ and $\alpha = 1/255$, keeping the number of steps consistent with the CIFAR setting. These parameter choices follow standard settings commonly used in adversarial training \cite{8835375}. To generate the common corruption training counterparts, the severity level is selected from the $1^{\text{st}}$–$3^{\text{rd}}$ range for each non-robust corruption type. This range introduces moderate perturbations that are sufficient to expose the model to corrupted inputs \cite{saikia2021improving}. For consistency, we applied the $3^{\text{rd}}$ severity level across all non-robust corruption types. 

Finally, to generate our validation and comparison datasets, we employed the default parameter settings for adversarial attacks as provided by the \texttt{torchattacks} package. For the corruption datasets, we utilized the severity level values for each corruption type as provided in \cite{hendrycks2018benchmarking}, ensuring consistency with established benchmarks.
\vspace{-5mm}
\subsection{Results and discussion}
\subsubsection{\textbf{Validation dataset comparison}}
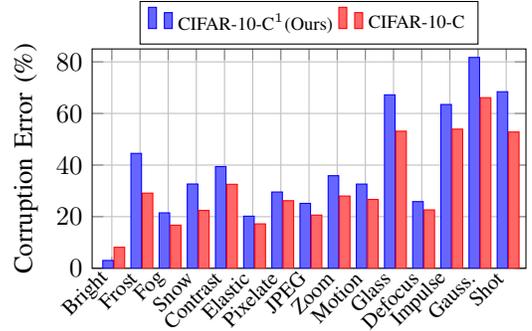
\begin{figure}[t]
\centering
\begin{tikzpicture}
\begin{axis}[
    ybar,
    bar width=4pt,
    enlarge x limits=0.05,
    symbolic x coords={Bright, Frost, Fog, Snow, Contrast, Elastic, Pixelate, JPEG, Zoom, Motion, Glass, Defocus, Impulse, Gauss., Shot},
    xtick=data,
    ylabel={Corruption Error (\%)},
    ymin=0, ymax=85,
    grid=both,
    xticklabel style={rotate=45, anchor=east, font=\footnotesize},
    legend style={at={(0.5,1.02)}, anchor=south, legend columns=-1, font=\scriptsize},
    width=0.85\columnwidth,
    height=4.5cm,
    xtick align=inside
]
\addplot+[fill=blue!60, bar shift=-0.05cm] coordinates {
    (Bright,3.034) (Frost,44.524) (Fog,21.486) (Snow,32.662) (Contrast,39.418) (Elastic,20.198) 
    (Pixelate,29.560) (JPEG,25.158) (Zoom,35.876) (Motion,32.634) (Glass,67.222) (Defocus,25.836) 
    (Impulse,63.474) (Gauss.,81.704) (Shot,68.414)
};
\addplot+[fill=red!60, bar shift=0.10cm] coordinates {
    (Bright,8.126) (Frost,29.124) (Fog,16.714) (Snow,22.418) (Contrast,32.552) (Elastic,17.218) 
    (Pixelate,26.198) (JPEG,20.620) (Zoom,28.004) (Motion,26.678) (Glass,53.122) (Defocus,22.638) 
    (Impulse,53.926) (Gauss.,66.110) (Shot,52.816)
};
\legend{CIFAR-10-C$^1$(Ours), CIFAR-10-C}
\end{axis}
\end{tikzpicture}
\caption{Performance comparison of AllConvNet on CIFAR-10-C\(^1\) and CIFAR-10-C. Our proposed validation dataset presents greater challenges across all corruption types except brightness.}
\label{fig-IV}
\end{figure}

We demonstrate that our proposed validation
datasets are ideal for immediate post-training evaluation of
standard models, in contrast to using perturbed holdout
test sets like CIFAR‑$10$-C and CIFAR‑$100$-C. Our findings indicate that although our validation set originates from the training data, when it is perturbed, it poses a more significant challenge to models than the perturbed holdout test set. This effect is especially pronounced for the non-robust perturbation types. For example, Figure \ref{fig-IV} compares AllConvNet’s performance on CIFAR‑$10$-C$^1$ versus CIFAR‑$10$-C. Except for brightness, all corruption types datasets in our validation dataset impair model performance more than those in CIFAR‑$10$-C. This supports our claim that evaluating models with weak robust samples magnifies the impact of challenging robustness scenarios. For instance, on the frost corruption in CIFAR‑$10$-C, the model’s error is below $\textbf{30\%}$, a level typically considered acceptable in robustness-critical applications \cite{hendrycks2018benchmarking, zhang2022understanding}. Whereas on CIFAR-$10$-C$^1$ the model is clearly non-robust to frost with an error of about $\textbf{45\%}$. We have observed the same trend especially for the non-robust perturbation types across all model architectures and datasets.
\vspace{-5mm}
\subsubsection{\textbf{Model Performance Evaluation}}
The relative adversarial error measurement of the models is not presented in this paper as all the considered adversarial perturbations fall in the non-robust group. Unlike the adversarial perturbations, for the corruption types, we chose those with the highest $\mathrm{RC_{Err}}$ values in bold and underlined for the corresponding standard models in Table \ref{tab:combined_rcerr}. For example, considering the standard AllConvNet in Table \ref{tab:combined_rcerr}, for the weather, digital, blur, and noise categories, the highest $\mathrm{RC_{Err}}$ values are from Frost, Contrast, Glass, and Gaussian for the respective categories. It can be observed and is evident that certain corruption types from each category consistently challenge the model's robustness across different architectures and datasets. To highlight the essence of our validation procedure, we measure the $\mathrm{RC_{Err}}$ of AugMix-enhanced models. We find that specific corruption types, such as Gaussian noise, still significantly degrade performance. For example, an AugMix-enhanced AllConvNet shows a $\mathrm{RC_{Err}}$ of $\textbf{7.50}$ on Gaussian-perturbed data. This represents a substantial deterioration relative to the clean baseline. This vulnerability arises because Gaussian noise disrupts both local and global features representations critical for classification. Even robustified strategies like AugMix cannot fully recover these features \cite{saikia2021improving}. These results show that, while enhancements improve overall robustness, certain corruptions remain inherently challenging. Therefore, it is important to explicitly include these difficult corruption types in augmentation strategies to further boost model performance \cite{GUO2023109308, gao2023out}. Consequently, we incorporate these challenging corruptions into the AugMix \cite{hendrycks*2020augmix} default augmentation operations. The corrupted versions of the training dataset is generated to boost the models' performance further.
\begin{table*}[t]
    \centering
    \caption{Relative Corruption Error ($\mathrm{RC_{Err}}$) of standard models on CIFAR-10-C\(^1\), CIFAR-100-C\(^1\), and ImageNet-100-C\(^1\). Highest $\mathrm{RC_{Err}}$ per corruption category is bold, underlined, and highlighted; these highlight the corruption types used for the targeted augmentation.}
    \fontsize{7.0}{9.0}\selectfont
    \setlength{\tabcolsep}{4pt} 
    \renewcommand{\arraystretch}{1.2} 
    \adjustbox{max width=\textwidth}{%
    \begin{tabular}{l|l|cccc|cccc|cccc|ccc}
        \toprule
        && \multicolumn{4}{c}{Weather} & \multicolumn{4}{c}{Digital} & \multicolumn{4}{c}{Blur} & \multicolumn{3}{c}{Noise} \\
        \toprule
        Dataset & Models (Err) & Bright & Frost & Fog & Snow & Contrast & Elastic & Pixelate & JPEG & Zoom & Motion & Glass & Defocus & Impulse & Gauss. & Shot \\
        \midrule
        \rowcolor{gray!10} 
        \multirow{3}{*}{CIFAR-10-C\(^1\)} 
        & AllConvNet & 0.465 & \cellcolor{blue!10}{\textbf{\underline{6.82}}} & 3.29 & 5.00 & \cellcolor{blue!10}{\textbf{\underline{6.04}}} & 3.09 & 4.53 & 3.85 & 5.50 & 5.00 & \cellcolor{blue!10}{\textbf{\underline{10.29}}} & 3.96 & 9.71 & \cellcolor{blue!10}{\textbf{\underline{12.52}}} & 10.47\\
        & DenseNet & 0.412 & \cellcolor{blue!10}{\textbf{\underline{7.81}}} & 2.80 & 5.71 & 5.35 & 4.19 & 6.20 & \cellcolor{blue!10}{\textbf{\underline{7.18}}} & 4.57 & 4.64 & \cellcolor{blue!10}{\textbf{\underline{10.74}}} & 3.46 & 9.50 & \cellcolor{blue!10}{\textbf{\underline{12.91}}} & 10.96\\
        \rowcolor{gray!10} 
        & WideResNet & 0.276 & \cellcolor{blue!10}{\textbf{\underline{6.77}}} & 2.28 & 5.45 & 4.22 & 3.68 & 5.99 & \cellcolor{blue!10}{\textbf{\underline{7.20}}} & 4.15 & 4.61 & \cellcolor{blue!10}{\textbf{\underline{10.77}}} & 2.43 & 8.44 & \cellcolor{blue!10}{\textbf{\underline{12.43}}} & 10.21\\
        \midrule
        \rowcolor{gray!10} 
        \multirow{3}{*}{CIFAR-100-C\(^1\)} 
        & AllConvNet & 0.437 & \cellcolor{blue!10}{\textbf{\underline{2.337}}} & 1.458 & 1.912 & \cellcolor{blue!10}{\textbf{\underline{2.085}}} & 1.306 & 1.725 & 1.929 & 1.826 & 1.814 & \cellcolor{blue!10}{\textbf{\underline{3.250}}} & 1.344 & 3.224 & \cellcolor{blue!10}{\textbf{\underline{3.584}}} & 3.318 \\
        & DenseNet & 0.394 & \cellcolor{blue!10}{\textbf{\underline{2.481}}} & 1.283 & 2.045 & 1.749 & 1.531 & 2.081 & \cellcolor{blue!10}{\textbf{\underline{2.454}}} & 1.870 & 1.843 & \cellcolor{blue!10}{\textbf{\underline{3.448}}} & 1.314 & 3.392 & \cellcolor{blue!10}{\textbf{\underline{3.761}}} & 3.441 \\
        \rowcolor{gray!10} 
        & WideResNet & 0.230 & \cellcolor{blue!10}{\textbf{\underline{2.674}}} & 1.721 & 2.063 & 2.043 & 1.894 & 1.944 & \cellcolor{blue!10}{\textbf{\underline{2.379}}} & 1.749 & 1.823 & \cellcolor{blue!10}{\textbf{\underline{3.448}}} & 1.171 & 3.219 & \cellcolor{blue!10}{\textbf{\underline{3.707}}} & 3.291 \\
        \midrule
        \rowcolor{gray!10} 
        \multirow{2}{*}{ImageNet-100-C\(^1\)} 
        & ResNet18  & 1.683 & 3.296 & 3.457 & \cellcolor{blue!10}{\textbf{\underline{3.635}}} & \cellcolor{blue!10}{\textbf{\underline{3.647}}} & 2.113 & 1.952 & 1.875 & 2.957 & 3.064 & 3.297 & \cellcolor{blue!10}{\textbf{\underline{3.308}}} & \cellcolor{blue!10}{\textbf{\underline{4.088}}} & 3.823 & 3.857 \\
        & Swin\_V2\_B  & 1.688 & 4.888 & 3.849 & \cellcolor{blue!10}{\textbf{\underline{5.405}}} & \cellcolor{blue!10}{\textbf{\underline{5.256}}} & 3.026 & 2.212 & 2.321 & 4.606 & 4.196 & \cellcolor{blue!10}{\textbf{\underline{5.332}}} & 4.823 & 5.075 & 4.978 & \cellcolor{blue!10}{\textbf{\underline{5.097}}} \\
        \bottomrule
    \end{tabular}}
    \label{tab:combined_rcerr}
\end{table*}
\begin{table*}[tbh]
\centering
\caption{Clean top-$1$-error rate (Err), Adversarial errors ($\mathrm{Adv_{Err}}$) across different attacks and mean adversarial error ($\mathrm{mAdv_{Err}}$) for various model architectures. Best results are in bold.}
\label{tab:adv_err_combined}

\begin{subtable}[t]{0.48\textwidth}
\centering
\caption{CIFAR-$10$-adv}
\label{tab:Table_7}
\adjustbox{max width=\textwidth}{%
\begin{tabular}{l c cccccc|c}
\toprule
\textbf{Network} & \textbf{Err}$\downarrow$ & BIM & FGSM & PGD & RFGSM & UAP & UMIFGSM & \bm{$\mathrm{mAdv_{Err}}$} \\
\midrule
AllConvNet (Std) & \(\mathbf{6.10}\) & 43.64 & 48.16 & 55.27 & 33.75 & 86.15 & 51.35 & 53.07 \\
AllConvNet (AugMix)   & 6.50 & 18.40 & 20.63 & 18.41 & 15.61 & 46.86 & 24.54 & 24.08 \\
AllConvNet (REVa) & \cellcolor{gray!10}6.31 & 
\cellcolor{gray!10}\textbf{8.67} & 
\cellcolor{gray!10}\textbf{10.94} & 
\cellcolor{gray!10}\textbf{8.28} & 
\cellcolor{gray!10}\textbf{8.03} & 
\cellcolor{gray!10}\textbf{26.11} & 
\cellcolor{gray!10}\textbf{10.89} & 
\cellcolor{gray!10}\textbf{12.15} \\
\midrule
DenseNet (Std) & 5.80 & 51.36 & 47.35 & 57.21 & 36.77 & 78.88 & 56.74 & 54.72 \\
DenseNet (AugMix)   & \textbf{4.90} & 17.70 & 18.40 & 16.15 & 13.66 & 44.82 & 23.84 & 22.43 \\
DenseNet (REVa) & \cellcolor{gray!10}5.05 & 
\cellcolor{gray!10}\textbf{8.90} & 
\cellcolor{gray!10}\textbf{11.20} & 
\cellcolor{gray!10}\textbf{8.06} & 
\cellcolor{gray!10}\textbf{8.10} & 
\cellcolor{gray!10}\textbf{18.04} & 
\cellcolor{gray!10}\textbf{11.08} & 
\cellcolor{gray!10}\textbf{10.90} \\
\midrule
WideResNet (Std) & 5.20 & 52.29 & 45.65 & 56.08 & 38.54 & 86.79 & 56.74 & 56.02 \\
WideResNet (AugMix)   & \textbf{4.90} & 16.52 & 18.67 & 14.74 & 13.41 & 35.86 & 22.27 & 20.14 \\
WideResNet (REVa) & \cellcolor{gray!10}5.16 & 
\cellcolor{gray!10}\textbf{8.67} & 
\cellcolor{gray!10}\textbf{11.80} & 
\cellcolor{gray!10}\textbf{8.09} & 
\cellcolor{gray!10}\textbf{8.07} & 
\cellcolor{gray!10}\textbf{15.43} & 
\cellcolor{gray!10}\textbf{11.31} & 
\cellcolor{gray!10}\textbf{10.56} \\
\bottomrule
\end{tabular}}
\end{subtable}
\hfill
\begin{subtable}[t]{0.48\textwidth}
\centering
\caption{CIFAR-$100$-adv}
\label{tab:Table_8}
\adjustbox{max width=\textwidth}{%
\begin{tabular}{l c cccccc|c}
\toprule
\textbf{Network} & \textbf{Err}$\downarrow$ & BIM & FGSM & PGD & RFGSM & UAP & UMIFGSM & \bm{$\mathrm{mAdv_{Err}}$} \\
\midrule
AllConvNet (Std) & 25.10 & 54.56 & 71.34 & 70.52 & 56.30 & 87.14 & 60.86 & 66.80 \\
AllConvNet (AugMix)   & 25.59 & 38.91 & 49.81 & 45.83 & 39.22 & 64.25 & 43.63 & 46.94 \\
AllConvNet (REVa) & \cellcolor{gray!10}\textbf{25.02} & 
\cellcolor{gray!10}\textbf{27.77} & 
\cellcolor{gray!10}\textbf{31.32} & 
\cellcolor{gray!10}\textbf{27.65} & 
\cellcolor{gray!10}\textbf{27.60} & 
\cellcolor{gray!10}\textbf{31.21} & 
\cellcolor{gray!10}\textbf{29.54} & 
\cellcolor{gray!10}\textbf{29.18} \\
\midrule
DenseNet (Std) & 28.60 & 57.95 & 69.31 & 69.92 & 58.52 & 86.50 & 63.54 & 67.62 \\
DenseNet (AugMix)   & 24.50 & 35.93 & 43.50 & 39.55 & 34.94 & 56.09 & 39.74 & 41.63 \\
DenseNet (REVa) & \cellcolor{gray!10}\textbf{22.78} & 
\cellcolor{gray!10}\textbf{27.80} & 
\cellcolor{gray!10}\textbf{31.14} & 
\cellcolor{gray!10}\textbf{27.97} & 
\cellcolor{gray!10}\textbf{27.43} & 
\cellcolor{gray!10}\textbf{31.41} & 
\cellcolor{gray!10}\textbf{29.58} & 
\cellcolor{gray!10}\textbf{29.22} \\
\midrule
WideResNet (Std) & 27.90 & 58.67 & 66.47 & 68.03 & 56.75 & 83.43 & 62.18 & 65.92 \\
WideResNet (AugMix)   & 24.20 & 34.09 & 41.53 & 38.18 & 34.48 & 52.18 & 38.92 & 39.90 \\
WideResNet (REVa) & \cellcolor{gray!10}\textbf{23.98} & 
\cellcolor{gray!10}\textbf{28.77} & 
\cellcolor{gray!10}\textbf{31.86} & 
\cellcolor{gray!10}\textbf{28.87} & 
\cellcolor{gray!10}\textbf{28.26} & 
\cellcolor{gray!10}\textbf{30.01} & 
\cellcolor{gray!10}\textbf{30.28} & 
\cellcolor{gray!10}\textbf{29.68} \\
\bottomrule
\end{tabular}}
\end{subtable}
\end{table*}
\begin{table}[tbh]
\centering
\captionsetup{skip=5pt}
\caption{Clean top-$1$-error rate (Err), Mean Corruption Error (mCE) comparison of standard, AugMix-enhanced, and REVa-enhanced models on CIFAR-10-C and CIFAR-100-C. Best results are in bold.}
\label{table_V}
\fontsize{6.5}{8.5}\selectfont
\setlength{\tabcolsep}{3pt}
\adjustbox{max width=\textwidth}{%
\begin{tabular}{@{}lllccc@{}}
\toprule
\textbf{Model} & \textbf{Method}  & \textbf{Err}$\downarrow$ & \textbf{CIFAR-10-C}$\downarrow$ & \textbf{Err}$\downarrow$ & \textbf{CIFAR-100-C}$\downarrow$ \\
\midrule
\multirow{3}{*}{AllConvNet}  
 & Standard & $\textbf{6.100}$ & $31.645$  & $25.10$ & $56.691$ \\
 & AugMix   & $6.500$  & $15.603$  & $25.59$ & $42.749$ \\
 & REVa (Ours) 
   & $6.310$ 
   & $\textbf{10.894}$ 
   & $\textbf{25.02}$ 
   & $\textbf{34.834}$\\ \midrule
\multirow{3}{*}{DenseNet} 
 & Standard & $5.800$ & $29.082$  & $28.60$ & $55.543$ \\
 & AugMix   & $\textbf{4.900}$ & $11.581$  & $24.50$ & $36.863$ \\
 & REVa (Ours) 
   & $5.050$ 
   & $\textbf{8.954}$  
   & $\textbf{22.78}$ 
   & $\textbf{33.041}$ \\ \midrule
\multirow{3}{*}{WideResNet} 
 & Standard & $5.200$ & $27.469$  & $27.90$ & $52.465$ \\
 & AugMix   & $\textbf{4.900}$ & $11.230$  & $24.20$ & $35.338$ \\
 & REVa (Ours) 
   &$5.160$ 
   & $\textbf{9.000}$  
   & $\textbf{23.98}$ 
   & $\textbf{32.573}$ \\
\bottomrule
\end{tabular}}
\end{table}
\subsubsection{\textbf{Model Robustness Enhancement}} 
 We present the comparison results of models enhanced via our enhancement procedure, \textbf{REVa}, with other methods. Our goal is to emphasize the importance of the validation procedure by demonstrating how its outcomes guided the robustness enhancement of the models on adversarial and common corruption types. Models enhanced with REVa match the clean test performance of both standard and AugMix-enhanced models while demonstrating superior robustness gains.
 Tables \ref{tab:Table_7} and \ref{tab:Table_8}  present the experimental comparison results of REVa-enhanced models with the standard and AugMix-enhanced models on CIFAR-$10$-adv and CIFAR-$100$-adv datasets respectively. The results are reported in terms of $\mathrm{Err}$, $\mathrm{Adv_{Err}}$, using  \eqref{eqn_3} and $\mathrm{mAdv_{Err}}$, using Equation \eqref{eqn_3E}. REVa-enhanced models achieve comparable performance on $\mathrm{Err}$ while attaining better performance as compared to models trained using the other methods. For instance, from Table \ref{tab:Table_7}, it can be observed that REVa-enhanced AllConvNet attains on average an error of $\textbf{12.153\%}$ as compared to the standard AllConvNet error of $\textbf{53.065\%}$ and AugMix-enhanced AllConvNet error of $\textbf{24.075\%}$ on the CIFAR-$10$-adv dataset. Similarly, from Table \ref{tab:Table_8}, we can observe that REVa-enhanced AllConvNet attains on average an error of $\textbf{29.182\%}$ on CIFAR-$100$-adv dataset as compared to adversarial error of $\textbf{66.795\%}$ and $\textbf{46.940\%}$ from standard AllConvNet and AugMix-enhanced AllConvNet respectively.
 Also, Table \ref{table_V} presents the comparison results of REVa-enhanced models with the standard and AugMix-enhanced models on CIFAR-$10$-C and CIFAR-$100$-C respectively. The results are reported in terms $\mathrm{Err}$ and $\mathrm{mCE}$ ( using Equation \eqref{eqn_4E}). Models enhanced using REVa, while achieving better performance on adversarial dataset, they achieve comparable or better performance than models enhanced via the AugMix technique on CIFAR-$10$-C and CIFAR-$10$-C datasets. From Table \ref{table_V} on CIFAR-$10$-C, we can observe that REVa-enhanced AllConvNet on average attains a corruption error  of $\textbf{10.894\%}$ as compared to standard AllConvNet and AugMix-enhanced AllConvNet corruption errors of $\textbf{31.645\%}$ and $\textbf{15.603\%}$ respectively. 
 
Similarly, from Table \ref{table_V} on the CIFAR-$100$-C, we can also observe that REVa-enhanced AllConvNet model  achieves on average a corruption error of $\textbf{34.834\%}$ as compared to standard AllConvNet and AugMix-enhanced AllConvNet corruption errors of $\textbf{56.691\%}$ and $\textbf{42.749\%}$  respectively.  The same observations as shown on Table \ref{table_V} are made across all of the models enhanced via REVa as they attain better performance on average compared to models trained via the other approaches on CIFAR-$10$ and CIFAR-$100$ datasets.

Table \ref{tab:table_VI} presents a performance comparison of the REVa-enhanced, standard, and AugMix-enhanced ResNet-$18$ as well as Swin\_V2\_B models on ImageNet-$100$-adv and ImageNet-$100$-C. The evaluation is in terms of $\mathrm{Err}$, $\mathrm{Adv_{Err}}$ and $\mathrm{mCE}$ respectively. On perturbed datasets, the REVa-enhanced ResNet-$18$ consistently outperforms both the standard and AugMix-enhanced models. It shows only a minimal degradation of $\textbf{0.1\%}$ in clean error. Meanwhile, adversarial error decreases by $\textbf{57.266\%}$ and corruption error by $\textbf{17.577\%}$ compared to the standard model. Against AugMix, REVa improves adversarial and corruption errors by $\textbf{12\%}$ and $\textbf{7\%}$, respectively. Similar gains are observed with a vision-based transformer model, highlighting the generality of REVa. We also report the statistical significance of the performance differences between REVa and the baseline methods. For each method, errors from different perturbation types (adversarial attacks or common corruptions) are grouped into a set. The Wilcoxon signed-rank test, with a significance level of $0.05$ is then applied to compare these sets. Table \ref{tab:table_stat_sig} presents the resulting p-values, reported to two decimal places for easy comparison with the significance level. Across all datasets, REVa is statistically significant than the Standard method. Compared with AugMix, REVa shows statistically significant improvements for most models, except for DenseNet and WideResNet on CIFAR-$100$-C and models evaluated on ImageNet-$100$-C. All p-values below $0.05$ are highlighted in bold, indicating that REVa achieves statistically significant improvements over the baseline Methods.\\
\begin{table}[htbp]
\centering
\captionsetup{skip=5pt}
\caption{Clean top-$1$-error rate (Err), Mean Adversarial Error ($\mathrm{mAdv_{Err}}$) and Mean Corruption Error ($\mathrm{mCE}$) comparison of standard, AugMix-enhanced, and REVa-enhanced models on ImageNet-100-adv and ImageNet-100-C. Best results are in bold.}
\label{tab:table_VI}
\fontsize{8}{9}\selectfont
\setlength{\tabcolsep}{3pt}

\begin{tabular}{@{}lllcc@{}}
\toprule
\textbf{Model} & \textbf{Method}  & \textbf{Err}$\downarrow$ & \textbf{mAdv\(_{Err}\)}$\downarrow$  & \textbf{mCE}$\downarrow$ \\
\midrule
\multirow{3}{*}{ResNet-18}  
 & Standard & \(18.540\) & \(93.366\) & \(56.956\) \\
 & AugMix   & \(\textbf{18.300}\) & \(47.972\) & \(46.536\) \\
 & REVa (Ours) 
   & \(18.640\) 
   & \(\textbf{36.100}\) 
   & \(\textbf{39.379}\) \\ \midrule
\multirow{3}{*}{Swin\_V2\_B} 
 & Standard & \(\textbf{7.200}\) & \(83.180\)  & \(30.123\) \\
 & AugMix   & \(8.720\) & \(55.603\) &\(27.543\) \\
 & REVa (Ours) 
   & \(8.480\) 
   & \(\textbf{22.356}\)  
   & \(\textbf{23.671}\) \\
\bottomrule
\end{tabular}
\end{table}

\begin{table}[htbp]
\centering
\captionsetup{skip=5pt}
\caption{Recorded p-values between the baseline methods and REVa. Cases where the p-value is less than the significance value ($0.05$) highlighted in bold, it shows that REVa is statistically significant than the compared approach.}
\label{tab:table_stat_sig}
\fontsize{8}{9}\selectfont
\setlength{\tabcolsep}{3pt}

\begin{tabular}{@{}lllc@{}}
\toprule
\textbf{Dataset} & \textbf{Model} & \shortstack{\textbf{Standard} \\ \textbf{\& REVa}} & \shortstack{\textbf{AugMix} \\ \textbf{\& REVa}}\\
\midrule
\multirow{3}{*}{CIFAR-$10$-adv}  
 & AllConvNet & $\textbf{0.03}$ & $\textbf{0.03}$  \\
 & DenseNet   & $\textbf{0.03}$ & $\textbf{0.03}$  \\
 & WideResNet & $\textbf{0.03}$ & $\textbf{0.03}$ \\ 
 \midrule
\multirow{3}{*}{CIFAR-$100$-adv} 
 & AllConvNet & $\textbf{0.03}$ & $\textbf{0.03}$  \\
 & DenseNet   & $\textbf{0.03}$ & $\textbf{0.03}$  \\
 & WideResNet & $\textbf{0.03}$ & $\textbf{0.03}$ \\ 
   \midrule
\multirow{2}{*}{ImageNet-$100$-adv} 
 & ResNet-$18$ & $\textbf{0.02}$ & $\textbf{0.03}$\\
 & Swin\_V2\_B   & $\textbf{0.00}$ & $\textbf{0.00}$\\
\midrule
\multirow{3}{*}{CIFAR-$10$-C} 
& AllConvNet & $\textbf{0.00}$ & $\textbf{0.00}$  \\
 & DenseNet   & $\textbf{0.00}$ & $\textbf{0.04}$  \\
 & WideResNet & $\textbf{0.00}$ & $\textbf{0.03}$ \\ 
\midrule
\multirow{3}{*}{CIFAR-$100$-C} 
& AllConvNet & $\textbf{0.00}$ & $\textbf{0.01}$  \\
 & DenseNet   & $\textbf{0.00}$ & $0.30$  \\
 & WideResNet & $\textbf{0.00}$ & $0.27$ \\ 
\midrule
\multirow{2}{*}{ImageNet-$100$-C} 
 & ResNet-$18$ & $\textbf{0.00}$ & $0.08$\\
 & Swin\_V2\_B  & $\textbf{0.03}$ & $0.14$\\
\bottomrule
\end{tabular}
\end{table}
\textbf{Effect of non-robust corruptions on the AugMix default augmentations:} 
Our goal is to demonstrate that targeted robustness enhancement not only improves performance on the specific perturbations it targets but also boosts performance across a broader range of perturbation types. We illustrate this by comparing models enhanced via REVa$^-$ and REVa. As defined in Section \ref{implement:details}, REVa$^-$ is identical to REVa, except that it uses the AugMix default augmentation operations to generate corruption counterparts from the original training dataset. We have observed that carefully incorporating corruption types from each category not only enhances performance on those specific corruptions \cite{saikia2021improving} but also improves performance on adversarial datasets. This underscores the importance of our validation procedure, which guides the selective integration of corruption types for targeted augmentation. Figures \ref{fig-additional_operations_cifar10_adv}--\ref{fig-additional_operations_cc} present the performance comparisons of AllConvNet models enhanced via REVa$^-$ and REVa, respectively.
Models enhanced using REVa consistently outperform those enhanced with REVa$^-$ on adversarial and common corruption datasets. For instance, Figures \ref{fig-additional_operations_cifar10_adv} and \ref{fig-additional_operations_adv} show that REVa-enhanced AllConvNet achieves superior results on the CIFAR-$10$-adv and CIFAR-$100$-adv datasets across all adversarial perturbation techniques. Similarly, Figure \ref{fig-additional_operations_cifar10_cc} demonstrates that on the CIFAR-$10$-C dataset, REVa-enhanced AllConvNet consistently outperforms its REVa$^-$ counterpart. For example, in the noise category, although AugMix’s default operations is updated with only Gaussian noise, further enhances performance on other noise corruptions. Similar improvements are observed for the other corruption types. In Figure \ref{fig-additional_operations_cc}, except motion blur and defocus blur, REVa-enhanced AllConvNet on CIFAR-$100$-C also outperforms REVa$^-$-enhanced AllConvNet across all corruption types. We hypothesize that adding glass blur to AugMix’s default operations may cause the model to focus on localized distortions, thereby reducing performance on motion and defocus blur. Similar observations on performance improvement were made for other models architectures on CIFAR-$10$, CIFAR-$100$ and ImageNet-$100$ datasets. Tables \ref{table_VI}  and \ref{table_VII} provide the comparison results for the other model architectures enhanced via REVa$^-$ and REVa. Models enhanced via REVa, on average, on the adversarial and common corruption datasets outperform REVa$^-$enhanced models. These performance gains underscore the power of our validation procedure. Focusing on non-robust perturbation types not only boosts performance on the targeted perturbations. It drives improvements across all perturbation types in the considered cross-domain settings. This demonstrates that our approach can be a valuable tool for refining and enhancing existing strategies.
\begin{figure}[htbp]
\centering
\begin{tikzpicture}
\begin{axis}[
    ybar,
    bar width=7pt,
    enlarge x limits=0.35,
    symbolic x coords={BIM, FGSM, PGD, RFGSM, UAP, UMIFGSM},
    xtick=data,
    ylabel={Adversarial Error (\%)},
    ymax=40,
    grid=both,
    xticklabel style={rotate=45, anchor=east, font=\footnotesize},
    legend style={at={(0.5,1.02)}, anchor=south, legend columns=-1, font=\scriptsize},
    width=0.85\columnwidth,
    height=4.5cm,
    xtick align=inside
]
\addplot+[fill=blue!60, bar shift=-0.05cm] coordinates {
    (BIM,12.18) (FGSM,16.39) (PGD,12.39) (RFGSM,10.91) (UAP,36.84) (UMIFGSM,17.06)
};
\addplot+[fill=red!50, bar shift=0.2cm] coordinates {
    (BIM,8.67) (FGSM,10.94) (PGD,8.28) (RFGSM,8.03) (UAP,26.11) (UMIFGSM,10.89)
};
\legend{REVa$^-$, REVa}
\end{axis}
\end{tikzpicture}
\caption[Short caption]{Adversarial error ($\mathrm{Adv_{Err}}$) comparison of REVa\(^-\)-enhanced and REVa-enhanced AllConvNet models on the CIFAR-$10$-adv dataset. The REVa-enhanced model shows better performance improvement.}
\label{fig-additional_operations_cifar10_adv}
\end{figure}
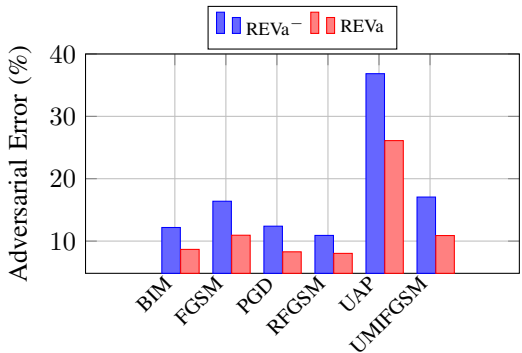
\begin{figure}[t]
\centering
\begin{tikzpicture}
\begin{axis}[
    ybar,
    bar width=7pt,
    enlarge x limits=0.35,
    symbolic x coords={BIM, FGSM, PGD, RFGSM, UAP, UMIFGSM},
    xtick=data,
    ylabel={Adversarial Error (\%)},
    ymax=40,
    grid=both,
     xticklabel style={rotate=45, anchor=east, font=\footnotesize},
    legend style={at={(0.5,1.02)}, anchor=south, legend columns=-1, font=\scriptsize},
    width=0.85\columnwidth,
    height=4.5cm,
    xtick align=inside
]
\addplot+[fill=blue!60, bar shift=-0.05cm] coordinates {
    (BIM,28.77) (FGSM,34.43) (PGD,29.96) (RFGSM,28.78) (UAP,38.26) (UMIFGSM,31.55)
};
\addplot+[fill=red!50, bar shift=0.2cm] coordinates {
    (BIM,27.35) (FGSM,30.82) (PGD,27.17) (RFGSM,27.27) (UAP,31.21) (UMIFGSM,29.05)
};
\legend{REVa$^-$, REVa}
\end{axis}
\end{tikzpicture}
\caption[Short caption]{Adversarial error ($\mathrm{Adv_{Err}}$) comparisons of REVa$^-$-enhanced and REVa-enhanced AllConvNet model on CIFAR-$100$-adv dataset. REVa-enhanced model shows better performance improvement.}
\label{fig-additional_operations_adv}
\end{figure}
\begin{figure}[htbp]
\centering
\begin{tikzpicture}
\begin{axis}[
    ybar,
    bar width=4pt,
    enlarge x limits=0.05,
    symbolic x coords={Bright, Frost, Fog, Snow, Contrast, Elastic, Pixelate, JPEG, Zoom, Motion, Glass, Defocus, Impulse, Gauss., Shot},
    xtick=data,
    ylabel={Corruption Error (\%)},
    ymax=25,
    grid=both,
     xticklabel style={rotate=45, anchor=east, font=\footnotesize},
    legend style={at={(0.5,1.02)}, anchor=south, legend columns=-1, font=\scriptsize},
    width=0.85\columnwidth,
    height=4.5cm,
    xtick align=inside
]
\addplot+[fill=blue!60, bar shift=-0.05cm] coordinates {
    (Bright,6.6) (Frost,12.502) (Fog,11.482) (Snow,12.172) (Contrast,23.664) (Elastic,10.428) 
    (Pixelate,12.502) (JPEG,11.734) (Zoom,9.464) (Motion,11.724) (Glass,24.03) (Defocus,8.26) 
    (Impulse,21.952) (Gauss.,20.096) (Shot,14.442)
};
\addplot+[fill=red!50, bar shift=0.10cm] coordinates {
    (Bright,6.456) (Frost,8.836) (Fog,10.878) (Snow,11.356) (Contrast,17.684) (Elastic,9.86) 
    (Pixelate,9.17) (JPEG,10.93) (Zoom,8.52) (Motion,11.5) (Glass,15.53) (Defocus,7.982) 
    (Impulse,14.234) (Gauss.,11.072) (Shot,9.4)
};
\legend{REVa$^-$, REVa}
\end{axis}
\end{tikzpicture}
\caption[Short caption]{Corruption error ($\mathrm{uCE}$) comparison of REVa\(^-\)-enhanced and REVa-enhanced AllConvNet models on the CIFAR-10-C dataset. REVa-enhanced model shows better performance across all corruption types.}
\label{fig-additional_operations_cifar10_cc}
\end{figure}
\begin{figure}[H]
\centering
\begin{tikzpicture}
\begin{axis}[
    ybar,
    bar width=4pt,
    enlarge x limits=0.05,
    symbolic x coords={Bright, Frost, Fog, Snow, Contrast, Elastic, Pixelate, JPEG, Zoom, Motion, Glass, Defocus, Impulse, Gauss., Shot},
    xtick=data,
    ylabel={Corruption Error (\%)},
    ymax=60,
    grid=both,
     xticklabel style={rotate=45, anchor=east, font=\footnotesize},
    legend style={at={(0.5,1.02)}, anchor=south, legend columns=-1, font=\scriptsize},
    width=0.85\columnwidth,
    height=4.5cm,
    xtick align=inside
]
\addplot+[fill=blue!60, bar shift=-0.05cm] coordinates {
    (Bright,28.81) (Frost,40.666) (Fog,38.388) (Snow,39.086) (Contrast,49.578) (Elastic,33.93) 
    (Pixelate,34.364) (JPEG,36.426) (Zoom,32.194) (Motion,34.117) (Glass,57.166) (Defocus,29.675)
    (Impulse,50.928) (Gauss.,48.12) (Shot,40.732)
};
\addplot+[fill=red!50, bar shift=0.10cm] coordinates {
    (Bright,27.86) (Frost,32.34) (Fog,36.142) (Snow,35.565) (Contrast,43.512) (Elastic,33.518) 
    (Pixelate,30.696) (JPEG,33.664) (Zoom,31.996) (Motion,35.626) (Glass,42.542) (Defocus,30.616)
    (Impulse,40.59) (Gauss.,34.762) (Shot,32.028)
};
\legend{REVa$^-$, REVa}
\end{axis}
\end{tikzpicture}
\caption[Short caption]{Corruption error ($\mathrm{uCE}$) comparison of REVa$^-$-enhanced and REVa-enhanced AllConvNet model on CIFAR-$100$-C dataset. REVa-enhanced model shows on average better performance across all corruption types.}
\label{fig-additional_operations_cc}
\end{figure}
\begin{table}[htbp]
\centering
\captionsetup{skip=5pt}
\caption{Mean Adversarial Error ($\mathrm{mAdv_{Err}}$) comparisons of REVa$^-$-enhanced and REVa-enhanced models on CIFAR-$10$-adv, CIFAR-100-adv and ImageNet-100-adv. Best results are in bold.}
\label{table_VI}
\fontsize{8}{9}\selectfont
\setlength{\tabcolsep}{3pt}
\begin{tabular}{@{}llcc@{}}
\toprule
& \textbf{Models} & \textbf{REVa$^-$}$\downarrow$ & \textbf{REVa}$\downarrow$ \\
\midrule
\multirow{2}{*}{CIFAR-$10$-adv}  
  & DenseNet  & 15.603  & \textbf{10.897} \\
  & WideResNet  & 17.877 & \textbf{10.562}\\ \midrule
\multirow{2}{*}{CIFAR-100-adv} 
  & DenseNet   & 42.414  & \textbf{29.223} \\
  & WideResNet  & 32.150  & \textbf{29.675}\\ \midrule
\multirow{1}{*}{ImageNet-100-adv} 
  & ResNet18   & 38.357  & \textbf{36.100} \\
  & Swin\_V2\_B   & 32.973  & \textbf{22.365}\\
\bottomrule
\end{tabular}
\end{table}
\begin{table}[htbp]
\centering
\captionsetup{skip=5pt}
\caption{Mean Corruption Error ($\mathrm{mCE}$) comparisons of REVa$^-$-enhanced and REVa-enhanced models on CIFAR-10-C, CIFAR-100-C and ImageNet-100-C. Best results are in bold.}
\label{table_VII}
\fontsize{8}{9}\selectfont
\setlength{\tabcolsep}{3pt}
\begin{tabular}{@{}lllcc@{}}
\toprule
& \textbf{Models} & \textbf{REVa$^-$}$\downarrow$ & \textbf{REVa}$\downarrow$ \\
\midrule
\multirow{2}{*}{CIFAR-10-C}  
 & DenseNet  & 15.603  & \textbf{8.954} \\
  & WideResNet  & 11.431 & \textbf{9.000}\\ \midrule
\multirow{2}{*}{CIFAR-100-C} 
 &  DenseNet & 37.463  & \textbf{33.041} \\
  & WideResNet & 35.864  & \textbf{32.573}\\ \midrule
\multirow{2}{*}{ImageNet-100-C} 
  & ResNet18 & 47.193  & \textbf{39.379} \\ 
  & Swin\_V2\_B & 28.675  & \textbf{23.671}\\
\bottomrule
\end{tabular}
\end{table}
\section{Conclusion}\label{section:5}
For robust data-driven models, it’s not enough to excel on clean data. We must expose and fix hidden vulnerabilities. In this paper, we introduced a validation procedure that both evaluates and boosts classifier performance under adversarial and common corruption conditions. Our approach creates validation datasets from “weak robust” samples. These are training data points that are sensitive to random perturbations. We identify such instances using local robustness analysis \cite{8103145}. Evaluating models on these challenging examples reveals vulnerabilities early on, complementing conventional robustness assessments. We then leveraged these insights to drive targeted improvements through data augmentation and a revised loss function, ultimately retraining the models to withstand both adversarial and common corruption. Our experiments on CIFAR and ImageNet-$100$ datasets show that our enhancement procedure (REVa) increases robustness, delivering superior performance on both adversarial and common corruption benchmarks.
\par In this work, we have demonstrated the effectiveness of our validation procedure considering a prominent enhancement technique, targeted enhancement. As a future direction, we plan to extend this validation procedure to evaluate its impact on generic augmentation approaches used in frequency-based enhancements as in \cite{saikia2021improving}. This will provide a broader understanding of the procedure's effectiveness across different model improvement techniques. \\
\vspace{-5mm}
\bibliographystyle{IEEEtran}
\bibliography{main}
\end{document}